\documentclass{article}

\usepackage[nonatbib,preprint]{neurips_2021}

\usepackage[utf8]{inputenc} 
\usepackage[T1]{fontenc}    
\usepackage{hyperref}       
\usepackage{url}            
\usepackage{booktabs}       
\usepackage{amsfonts}       
\usepackage{nicefrac}       
\usepackage{microtype}      
\usepackage{xcolor}         
\usepackage{graphicx}
\usepackage{algorithm}
\usepackage{algpseudocode}
\usepackage{bbm}
\usepackage{biblatex}
\usepackage{subfigure}
\title{Dual reparametrized Variational Generative Model for Time-Series Forecasting}

\author{
  Ziang Chen \\
  Department of Engineering\\
  King's College London\\
  Strand, London, WC2R 2LS \\
  \texttt{ziang.chen@kcl.ac.uk} \\
}

\begin{document}

\maketitle

\begin{abstract}
 This paper propose DualVDT, a  generative model for Time-series forecasting. Introduced dual reparametrized variational mechanisms on variational autoencoder (VAE) to tighter the evidence lower bound (ELBO) of the model, prove the advance performance analytically. This mechanism leverage the latent score based generative model (SGM), explicitly denoising the perturbation accumulated on latent vector through reverse time stochastic differential equation and variational ancestral sampling. 
The posterior of denoised latent distribution fused with dual reparametrized variational density. The KL divergence in ELBO will reduce to reach the better results of the model. This paper also proposed a latent attention mechanisms to extract multivariate dependency explicitly. Build the local-temporal dependency simultaneously in factor wised through constructed local topology and temporal wised. The proven and experiment on multiple datasets illustrate, DualVDT, with a novel dual reparametrized structure, which denoise the latent perturbation through the reverse dynamics combining local-temporal inference, has the advanced performance both analytically and experimentally.
\end{abstract}

\section{Introduction}


Multivariate Time Series forecasting, which extensively applied  in multiple disciplines like fiance\cite{tf_review1}, economy\cite{tf_dl_review}, epidemic\cite{chen_epidimic}, self-driving \cite{human_trajectory}, etc., have caught enormously research which summarized in \cite{tf_review1}\cite{tf_dl_review}. The most advanced results are mainly implemented leverage Deep Generative models, includes  normalized flows \cite{nf_review}, variational autoencoder (VAE) \cite{vae}\cite{intro_vae}, generative adversarial network (GAN) \cite{gan} etc. \cite{generative_models}\cite{semi_generative}\cite{auxilary}. Among them, VAE using variational inference,  can captured the importance factor and give the density estimation, \cite{nvae}\cite{vae} demonstrate the advance of VAE  compared to  other  generative models. Additionally, the experiment on  multiple tasks shows impressive results of VAE\cite{gpvae}\cite{som_vae} \cite{deep_tf} .

However, main issue encounter during temporal inference of VAE,  is the variation lower bound may divergence with time increased \cite{temporal_vae}\cite{nf_review}, i.e. the error, which  closely related to dependency formulation, will accumulated in such autoregressive process, thus restrict the real-word performance of the models \cite{vanishing}. To build a applicable dependency of multivariate time series. It can be categorized to factor to factor and temporal to factor (also called spatial-temporal interaction) phases \cite{human_trajectory}\cite{tf_dl_review}. Methods in factor to factor interaction where each elements may affect each other at various level, can be captured explicitly with a topological structure in graph neural network (GNN)\cite{egnn}\cite{hgnn}, or implicitly by the local operation as pooling or convolution \cite{clstm}. On temporal dependency, same method in implicit local operation can be applied directly \cite{clstm} or combining  temporal information. One may make a natural assumption of the importance decay with the recall steps increased \cite{lstm_review} or establish the temporal dependency separately compared to embedding in the hidden state through exploit the self-attention mechanism\cite{attention_all_you_need}\cite{informer}.

Although the methods above illustrate the advance in practice \cite{informer}\cite{tf_dl_review}. On theoretical perspective, scored based generative models (SGM), explicit denoising the perturbation on latent space has demonstrated the impressive result on audio and image generation \cite{lsgm}\cite{song_score}\cite{wavegrad}. Latent SGM have forward and reverse process. At forward process, it defined a noise diffusion process on latent space by stochastic differential equation (SDE), then learning the log-density of perturbation, i.e., the distribution of noise injection\cite{song_score}\cite{lsgm}. To reduce the noise, it using learned score model in reverse time SDE, which can been seen as ancestral sampling from perturbed density to  'pure density' \cite{denoising}. This approach thus will optimized the evidence lower bound of the variational probabilistic model. 

\begin{figure}[H]
  \centering
  \includegraphics[width=12.5cm]{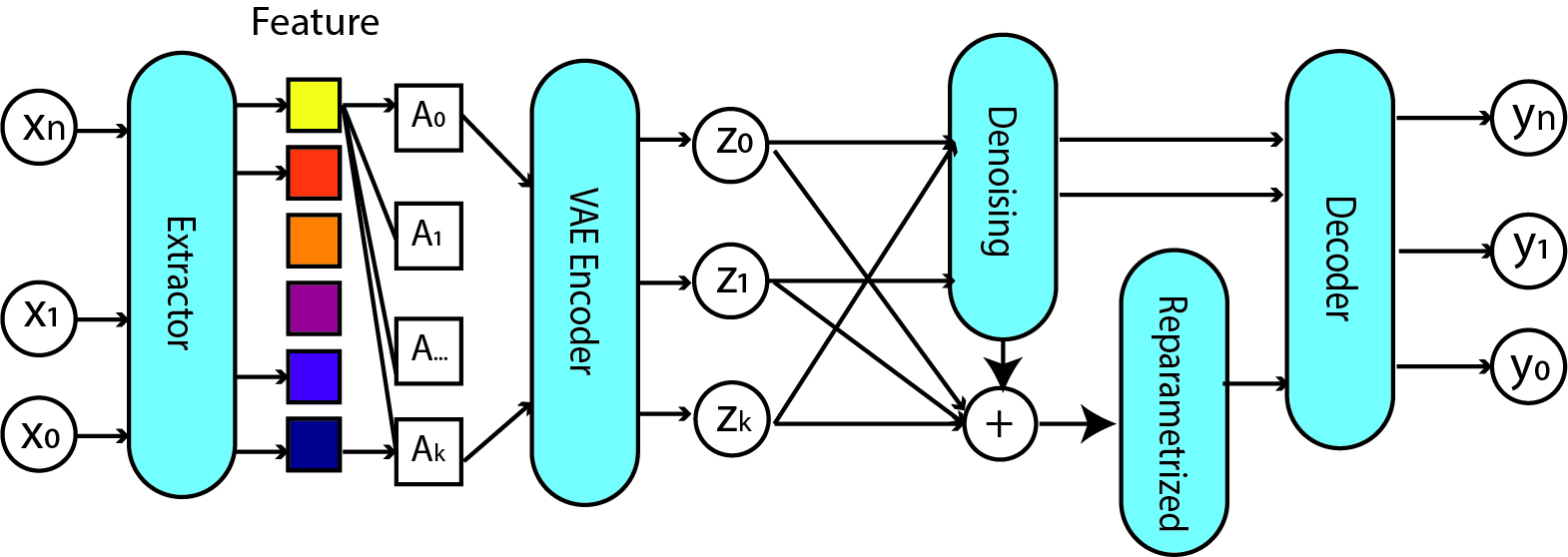}
  \caption{Structure of DualVDT, multivariate time-series data first feed in to the encoder, extract the feature accordingly and build their local-temporal dependency. Then pass to the Dual reparametrized on reverse SDE phase to tighter the variational lower bound of the model}
  \label{structure}
\end{figure}

This paper leverage both advance in practice and analytical, introduced DualVDT, a model which encompass the latent score based generative model and local-temporal aggregation operator into multivariate time series forecasting problem. Leverage the learned dependency with a novel dual reparametrized mechanism. A tighter variation lower bound thus can be construct and proven tighter than original VAE. i.e., ensure the  better performance in  theoretically. DualVDT also exploit self-attention mechanism on dynamic topological, inference the spatial temporal interaction simultaneously.  On behalf of the experiment, this paper evaluate on ETDataset (Electricity Transformer Dataset) \cite{informer} \footnote{https://
github.com/zhouhaoyi/ETDataset} and Covid-19 dataset \cite{covid19} \footnote{https://health.google.com/covid-19/open-data/raw-data}.Summarized main contributions in this paper are:

\paragraph{Dynamic Local Temporal Dependency on Multivariate Time Series :}  This model explicit learn the dynamic dependency in multivariate time series data on local and temporal interaction. Introduce a local-temporal inference mechanism which can simultaneously exploit both types of interaction with high efficiency.

\paragraph{Dual Reparametrized Variational inference :} DualVDT introduce a dual reparametrized variational inference on the latent space. Through the latent score based generative model, which explicitly denoise the perturbation accumulated by reverse time stochastic differential equation.

\paragraph{Tighter Evidence Lower Bound} With dual reparametrized variational, the evidence lower bound can be proven tighter than original VAE, ensure the performance of the model. The ablation study also illustrate the effect of this mechanism.

\section{Background}

\subsection{Score Based Generative Model}


This section review the score-based generative models which reduced the perturb through matching the score function in diffusion process, directly estimate the noise injection. Considered a random vector, in this paper, represent the latent code $\mathbf{z}$ defined by the encode process $p_\theta(\mathbf{z}|\mathbf{x})$ (discussion of latent space in section 2.2). A perturbation process on latent space at time $t$ has the transition density $q(\mathbf{z}_t|\mathbf{z}_0)$. Described by Stochastic Differential Equation (SDE) below. 

\begin{equation}
\mathrm{d} \mathbf{z}= f(t) \mathbf{z} +g(t) \mathrm{d}\mathbf{w}
\label{forward}
\end{equation}

(\ref{forward}) is the process maps the latent code between two same $k$ dimensional space and inject the noise, where $\mathbf{w}$ is the standard Wiener process, $f(t)$ is the model drift, $g(t)$ is the diffusion, satisfied $f(t):\mathbb{R}^k \to \mathbb{R}^k, g(t):\mathbb{R}^k \to \mathbb{R}^k$. Through this process, density will converge to Gaussian when time tend to infinity, represent the error accumulation in the auto-regressive model. Follow \cite{song_score}, the reverse-time SDE desrible  $q(\mathbf{z}_0|\mathbf{z}_t)$ is :


\begin{equation}
\mathrm{d} \mathbf{z}=\left[f(t) \mathbf{z}-g(t)^{2} \nabla_{\mathbf{z}} \log q_{t}(\mathbf{z})\right] \mathrm{d} t+g(t) \mathrm{d} \overline{\mathbf{w}}
\label{reverse}    
\end{equation}

where $\overline{\mathbf{w}}$ is the reverse Wiener process, the goal of the SGM is to train a score function which estimate the score $\nabla_{\mathbf{z}} \log q_{t}(\mathbf{z})$ to reverse the diffusion process on arbitary density of random vector $\textbf{z}$ through the ancestral sampling of reverse transition $q(\mathbf{z}_0|\mathbf{z}_t)$. The prior density $q(\mathbf{z})$ can be estimated via score matching objective:

\begin{equation}
\min _{\theta} \mathbb{E}_{t}\left[\lambda(t) \mathbb{E}_{q\left(\mathbf{z}_{0}\right)} \mathbb{E}_{q\left(\mathbf{z}_{t} \mid \mathbf{z}_{0}\right)}\left[\left\|\nabla_{\mathbf{z}_{t}} \log q\left(\mathbf{z}_{t} | \mathbf{z}_{0} \right)-\nabla_{\mathbf{z}_{t}} \log p_{\theta}\left(\mathbf{z}_{t}\right)\right\|_{2}^{2}\right]\right] + C
\label{score_matching}
\end{equation}


Coefficient $C$ and weighting $\lambda(t)=\frac{1}{2}g(t)^2$ are parameter independent referenced to\cite{lsgm}

\begin{equation}
C=\mathbb{E}_{t}\left[\lambda(t) \mathbb{E}_{q\left(\mathbf{z}_{0}\right)} \mathbb{E}_{q\left(\mathbf{z}_{t} \mid \mathbf{z}_{0}\right)}\left[\left\|\nabla_{\mathbf{z}_{t}} \log q\left(\mathbf{z}_{t}\right)\right\|_{2}^{2}-\left\|\nabla_{\mathbf{z}_{t}} \log q\left(\mathbf{z}_{t} \mid \mathbf{z}_{0}\right)\right\|_{2}^{2}\right]\right]
\end{equation}

\subsection{SGM in Latent Space}


\cite{lsgm} introduced latent space with a diffusion denoising process above. The common prior of latent $p(\mathbf{z}_0)$ in VAE \cite{vae}\cite{nvae} is gaussian $\mathcal{N}(\mathbf{z}_0;0, \sigma_{0}^{2}\mathbf{I})$, the diffused density  $q\left(\mathbf{z}_{t} \mid \mathbf{z}_{0}\right)=\mathcal{N}\left(\mathbf{z}_{t} ; \mu_{t}\left(\mathbf{z}_{0}\right), \sigma_{t}^{2} \mathbf{I}\right)
$ According to \cite{lsgm}, a latent space generated by the encoder $q_\phi(\mathbf{z}|\mathbf{x})$ with a score matching prior $p_\theta(\mathbf{z})$, the cross entropy loss is:

\begin{equation}
\mathcal{L}\left(q_{\phi}\left(\mathbf{z}_{0} \mid \mathbf{x}\right)|| p_{\theta}\left(\mathbf{z}_{0}\right)\right)=\mathbb{E}_{t }\left[\frac{w(t)}{2} \mathbb{E}_{q_{\phi}\left(\mathbf{z}_{t}, \mathbf{z}_{0} \mid \mathbf{x}\right), \epsilon}\left[\left\|\epsilon-S_{\theta}\left(\mathbf{z}_{t}, t\right)\right\|_{2}^{2}\right]\right]+\frac{k}{2} \log \left(2 \pi e \sigma_{0}^{2}\right)
\label{ce_loss}    
\end{equation}

The score model $S_{\theta}$ which defined in section 3, is the approximation of the score function in (\ref{reverse}), $\epsilon$ is the perturbation in diffused sampling $\mathbf{z}_t \sim q\left(\mathbf{z}_{t} \mid \mathbf{z}_{0}\right)$ , parameterized by $ \mathbf{z}_t=\mu_{t}\left(\mathbf{z}_{0}\right)+\sigma_t\epsilon$. Time-depend weight scale coefficient $w(t)=\frac{g(t)^2}{\sigma_t^2}$.  
The goal of latent score-based model is minimize both reconstruction error in VAE \cite{vae} and score matching error \cite{lsgm} with the combined loss.

\begin{equation}
Loss = \mathcal{L}\left(q_{\phi}\left(\mathbf{z}_{0} \mid \mathbf{x}\right)|| p_{\theta}\left(\mathbf{z}_{0}\right)\right) +  \mathcal{L}\left(q_{\phi}\left(\mathbf{z}_{0} \mid \mathbf{x}\right)|| p_{\psi}\left( \mathbf{x} \mid \mathbf{z}_{0}  \right)\right) 
\end{equation}

Where $p_{\psi}\left( \mathbf{x}  \mid \mathbf{z}_{0}   \right)$ is the decoder. The joint distribution thus can be written as $p(\mathbf{z}_0,\mathbf{x})=p_\theta(\mathbf{z})p_{\psi}\left( \mathbf{x}  \mid \mathbf{z}_{0}   \right)$.


\section{Multivariate Temporal Generative Model}



\subsection{Local Temporal Inference}

The structure of DualVDT as Figure \ref{structure} shows. This section will firstly, formulate a multivariate time series forecasting inference problem. Then introduce the local temporal neighbourhood aggregation operator. Consider a group of series $\textbf{x}=(\textbf{x}_0, \textbf{x}_1, ... ,\textbf{x}_{n_x})^T$ which have $n_x$ variables. Each variable has a sequence of history point $\mathbf{x}_i = (x_{i0},x_{i1},...x_{iT_x}) $ within a given look back window  $t = 0,1,....T_x$. Foretasted series $\textbf{y}=(\textbf{y}_0, \textbf{y}_1, ... ,\textbf{y}_{n_y})^T$, $\mathbf{y}_i = (y_{i0},y_{i1},..., y_{iT_y}) $ with $n_y$ series in total time steps $T_y$. The forecasting process in probabilistic can be viewed as sampling from learned posterior distribution $\textbf{y} \sim p_\psi( \textbf{y} \mid \textbf{x} )$. For the convenient of discussion, let $n_x,n_y$ and $T_x,T_y$ equal to each other and denote as $n, T$ in later text, extend of the unmatched dimension can simply via padding zeros on the smaller dimension and masked out the useless feature as \cite{attention_all_you_need}. 

To speed up the computation and reach spatial temporal aggregation simultaneously, reference to \cite{agentformer}, define a local-temporal mask:

\begin{equation}
\gamma_{ab} = \mathbbm{1}  \left( a \, mod \, n = b \, mode \, n \right) 
\end{equation}

\begin{figure}[H]
  \centering
  \includegraphics[width=5cm]{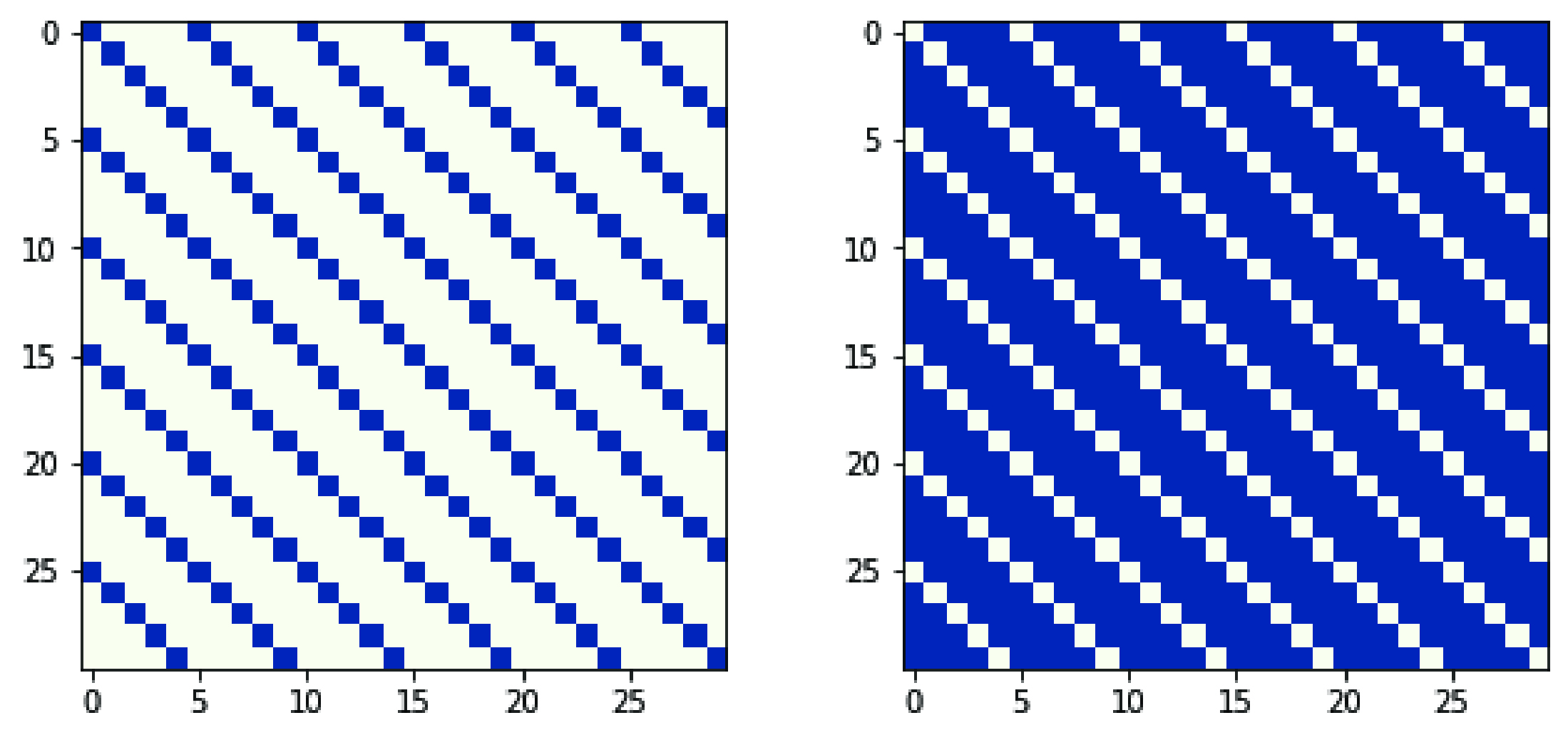}
  \caption {An example of the temporal mask  $\gamma$ and local mask $1-\gamma$} 

\end{figure}

Applying the multi-head self intention mechanism \cite{attention_all_you_need}, the inference with local temporal masks can evaluated simultaneously with learned query, key and value by extractor $Q(\textbf{x}), K(\textbf{x}), V(\textbf{x})$, details will given at Section 4.

\begin{equation}
Att(\mathcal{W},\textbf{x})=Softmax(\frac{ w^q Q(\textbf{x}) \times w^k K(\textbf{x}) }{n})w^v V(\textbf{x})
\label{att}
\end{equation}

The weights of query, key and value tensor $\mathcal{W}=(w^q,w^k,w^v)$, aggregate the attention and mask, the multi-level local-temporal dependency:

\begin{equation}
 \mathcal{A} =self-Att_\gamma ( \textbf{x}) = Att( \mathcal{W}_{\alpha},\gamma \odot \textbf{x})+  Att(\mathcal{W}_{\beta}, (1-\gamma) \odot \textbf{x})
 \label {attn}
\end{equation}

Where $\odot$ is the elementwise product, $\mathcal{W}_{\alpha}, \mathcal{W}_{\beta}$ are weights tensor of local and temporal correspondingly. Above dependency built by multi-head self attention can be seen as a dynamic topological on heterogeneous graph, with two types of link, the spatial (\ref{local_edge}) and temporal (\ref{temporal_edge}), in heterogeneously graph \cite{HGT}:

\begin{equation}
[e^l] {i , j} =   Att( \mathcal{W}_{\alpha},\gamma \odot \textbf{x})
\label{local_edge}
\end{equation}

\begin{equation}
[e^t]_{t_1,t_2}=Att(\mathcal{W}_{\beta}, (1-\gamma) \odot \textbf{x})
\label{temporal_edge}
\end{equation}

The node represents the factors embedding with the same way in (\ref{att}) 

\subsection{Dual Reparametrized Variational}

After built the local-temporal dependency, encoding the latent with the dual reparametrized variational mechanism. Let a posterior of latent vector $\textbf{z}$ as latent $p_{\theta}\left(\mathbf{z}\right)$ with parameter $\theta$, the denoising transition density $q\left(\mathbf{z}_{t} \mid \mathbf{z}_{0}\right)$ and reverse transition $q(\mathbf{z}_0|\mathbf{z}_t)$ can be obtained through (\ref{reverse}). The variance convergence according:

\begin{equation}
0<...\sigma_{t-1}^{2}<\sigma_{t}^{2}<1
\label{convergence}
\end{equation}

Scaled the exception through $\mu_t(\textbf{z}_{t-1})=\sqrt{1-\sigma_t^2}\textbf{z}_{i-1}$. The training goal of the optimal posterior score density parameter $\theta$  in (\ref{ce_loss}) re-weighted to:
\begin{equation}
\theta^{*} =argmin_\theta  \sum_{i=1}^{N}\left(1-\alpha_{i}\right) \mathbb{E}_{p_{\theta}(\mathbf{z})} \mathbb{E}_{p_{\alpha_{i}}(\textbf{z}_t} \mid \mathbf{z})\left[\left\|\mathbf{s}_\theta({\mathbf{z}}_t, i)-\nabla_{\mathbf{z}_t} \log p_{\alpha_{i}}(\mathbf{z}_t \mid \mathbf{z})\right\|_{2}^{2}\right]
\label{elbo_as}
\end{equation}

Where $\alpha_i=\prod_{t=0}^i \sigma_t^2$,  (\ref{elbo_as}) defined ancestral sampling of the reverse transition density with:

\begin{equation}
q(\mathbf{z}_{t-1}|\mathbf{z}_t) =\mathcal{N}\left(\mathbf{z}_{t-1} ; \frac{1}{\sqrt{1-\sigma_t^2}}(\textbf{z}_i + \sigma_t^2s_\theta(\textbf{z}_t,t) , \sigma_{t}^{2} \mathbf{I}\right) 
\label{as}
\end{equation}

With a sequence of sampling, $p_{\theta}(\mathbf{z}_{t-1})= q(\mathbf{z}_{t-1}|\mathbf{z}_t) p_{\theta}(\mathbf{z}_{t-1})$, the denoised latent $\textbf{z}_0$ sampled with a dual reparameterized through

\begin{equation}
\textbf{z} \sim  \mathcal{N}\left(\textbf{z}_0 ; \mu_f( z_0^\theta + z_0^\phi) ,\frac{1}{\sqrt{1-\sigma_t^2}}  \Sigma( z_0^\theta + z_0^\phi  ) \right)
\label{dual_rr}
\end{equation}

The sampling $\textbf{z}_0^\theta  \sim p_{\theta}(\textbf{z}_0) $ , $\textbf{z}_0^\phi  \sim p_{\phi}(\textbf{z}_0 \mid \textbf{x}) $, the score mathcing objective therefor becomes maximize the likelihood in dual reparameterized process. Score function $s_\theta$ can be arbitrary model with a refineable parameter between same dimensional space as  (\ref{reverse}).

\subsection{DualVDT and Variational Lower Bound}

The model introduced in this paper, has the process shown in Figure \ref{structure} and Algorithm 1.  During inference, the multivariate temporal data formulated in Section 3.1. Factor interaction in local and temporal through learned dynamic dependency in (\ref{att}), $\gets_s$ is the correspondingly sampling method in variational inference \cite{vae} and in (\ref{as}). Latent  feed into encoder and the reverse phase in SGM on its latent space optimized as (\ref{elbo_as}). With a sequence of sampling, the denoised posterior of latent transformed with dual reparameterized variational mechanism in (\ref{dual_rr}), denote as $\mathcal{D}$  ensure the conjunct of prior in (\ref{ce_loss}). This paper let the density estimation between reconstruction decoder and future predication identity with $p_\psi$, thus DualVDT have the ability evaluate both predication and imputation with different mask.  

\begin{algorithm}[H]
\centering
\caption{DualVDT}\label{alg_dualvdt}
\begin{algorithmic}
\State $X \gets [x_i], i=0,1,...n$
\State $\gamma \gets [\mathbbm{1}  \left( a \, mod \, n = b \, mode \, n \right) ]_{ab}$
\While{$i  \gets 0,..N$}

	$A \gets self-Att_\gamma (X)$
	
    $\textbf{z}_t \gets_S  q_\phi (\textbf{z} \mid \textbf{x})=\mathcal{N}(\mu(A) , \sigma(A)^2\mathbf{I})  $
	\While {$j  \gets 1,..t$}
	
    		$\textbf{z}_{t-j} \gets_S  p_\psi (\textbf{z}_{t-j} \mid \textbf{z}_{t-j+1} ) $
	\EndWhile
    
    $\textbf{z}_0 \gets_S \mathcal{N}(\mu(\textbf{z}_0) , \sigma(\textbf{z}_0)^2\mathbf{I}) + \textbf{z}_t $
    
    $\textbf{y} \gets \mathcal{D}(p_\psi ( \textbf{y} \mid \textbf{z}_0 ))$
\EndWhile
\end{algorithmic}
\end{algorithm}

The variational lower bound of DualVDT can be prove tighter compared to the original ELBO of VAE. For vanilla VAE which has the ELBO as \cite{vae}:

\begin{equation}
\mathcal{L}(\phi) =\mathbb{E}_{\mathbf{z} \sim q}\left[\log p(\mathbf{x} \mid \mathbf{z})+log p(\mathbf{z})-\log q_{\phi}(\mathbf{z} \mid \mathbf{x})\right]   
\end{equation}

Spilt the last two terms , a KL divergence between latent density and posterior can be separate

\begin{equation}
\mathcal{L}(\phi)  = \mathbb{E}_{\mathbf{z} \sim q}[\log p(\mathbf{x} \mid \mathbf{z})]-K L\left(q_{\phi}(\mathbf{z} \mid \mathbf{x}) \| p(\mathbf{z})\right) 
\end{equation}

Since the latent generative model estimate the posterior distribution of the latent, reduce KL divergence in at score based generative shown in (\ref{ce_loss}), the convergence in (\ref{convergence}) will ensure the convergence of such algorithm into a tighter ELBO\footnote{due to the basic relationship between KL divergence and cross entropy $KL(p||q) = H(p,q) - H(p,p)$}.

\section{Experiments}

On behalf of the experiments, this paper first compare different sequence to sequence model as Table \ref{compartion}, then evaluate the ablation study to illustrate the convergence tighter than original VAE as Table \ref{ablation}. Training on datasets ETDataset\cite{informer} and Covid-19 open data \cite{covid19} asses the mean square error (MSE) and mean absolute error (MAE).

\begin{table}[H]
	\small
    \centering
    \caption{Multivariate Time Series forecasting of different model}
    \begin{tabular}[H]{c|c|c|c|c}
      & \multicolumn{2}{|c|}{ETTH}  & \multicolumn{2}{c}{Covid-19}  \\
     \hline
      Model &  Mean Square Error  & Mean Absolute Error &
      Mean Square Error  & Mean Absolute Error  \\     
      \hline
      LSTM & 0.955 & 0.708  & 0.534 & 0.425 \\
      Transformer &  0.621  & \textbf{ 0.239}  &  0.583 &  0.221 \\
        Gaussian Process  & 1.590   &  1.019  &  1.870  &  1.437  \\
      DualVDT  &  \textbf{0.422}  & 0.438 &  \textbf{0.370} & \textbf{0.209}  
    \end{tabular}
    \label{compartion}
\end{table}

In ETDataset, the factor include High Useless Load (HUL), Middle UseFul Load (MUFL), Low UseFul Load (LUFL) and Low UseLess Load (LULL), the target varibale is Oil Temperature (OT).  According to \cite{informer}, to predict the usage varies with the time based on the factor HUL, MUFL LUFL and LULL, due to OT is necessary and which can reflect he condition of Electrical Transformer (ET), thus choice as the target variable.  

Covid-19 Open Data is the epidemiological database on global. The pandemic problem has the nature property of the topology interaction. Infected population migrate to effect other region, this paper using the same approach in \cite{chen_epidimic} to divide the dataset and doing the validation.

Using the same setting of parameter as in \cite{lstm_review}\cite{attention_all_you_need}\cite{nf_review}. The results shows in most cases, DualVDT can performer better than other sequence models in these dataset. One reason may because LSTM and Transformer are build for classify output (of the word embedding) rather than regression. And Gaussian Process which can been seen as the DualVDT with linear encoder decoder and remove both dual reparamenterized and local-temporal inference.

\begin{table}[H]
	\caption{Ablation Study on DualVDT}
    \small
    \centering
\begin{tabular}[H]{c |c | c |c | c | c  | c |c  }

\multicolumn{4}{c}{ DualVDT Component}	   & \multicolumn{2}{|c|}{ETTH}  & \multicolumn{2}{c}{Covid-19} 	\\
\hline
Encoder & Score Model & Dual Reparametrized  & Latent Sampler &  MSE & MAE &  MSE & MAE  \\
\hline
FC   & FC  &   OFF  &  AS   & 0.860  &  0.943  &   0.948 &  0.792 \\ 
FC   & FC  &   OFF  &  RD   & 0.941  &   0.912  &  0.840 & 0.996 \\ 
FC   & FC  &   OFF  &  PF   & 0.905  & 1.152  &  0.624  &  0.706 \\ 
LT   & FC  &   OFF  &  AS   & 0.723  &  0.701 & 0.713 & 0.907 \\
FC   & CNN  &  OFF  &  AS   & 0.725  &  0.880 & 0.791  &  0.652 \\
FC   & FC  &   ON  &  AS    & 0.638  &  0.661 & 0.452  &  0.477 \\ 
LT   & FC  &   ON  &  AS    & 0.597  &  0.542  & \textbf{0.305}  &  0.298 \\ 
FC   & CNN  &  ON  &  AS    & 0.542  &  0.667  & 0.455  &  0.305 \\ 
LT   & CNN  &  ON  &  AS    & \textbf{0.422}  & \textbf{0.438}   & 0.370  &  \textbf{0.209}\\ 
\end{tabular}
    \label{ablation}
\end{table}    
    
On ablation study, this paper select different setting of encoder, score model with various sampling method introduced by \cite{song_score}. The decoder has the same structure with encoder except the output dimension matching the dimension of target variable. In Table \ref{ablation}, asses Fully Connect (FC), Convolution Neural Network (CNN) and Local Temporal Attention (LT) on different stage as Figure \ref{structure} shows. The sampling method includes ancestral sampling, reverse differential equation solved by NeuralODE, and probability flow method \cite{song_score}. 

The results demonstrate as the prove in Section 3. Dual Reparametrized can reduce the loss of the model. In most cases, local temporal attention have more accurate estimation than vanilla VAE using Multi-layer perceptions. The assessment of different sampling method also illustrate the same results as \cite{song_score}.


\section {Conclusion}
This paper propose DualVDT, a  generative model for Time-series forecasting with dual reparametrized variational mechanisms. The model can be proven have a tighter evidence lower bound (ELBO) which ensure the performance compared to vanilla VAE. With score based generation model, reduce the KL divergence through score matching process explicitly by reverse the perturbation process. This paper also proposed a latent attention mechanisms to extract multivariate dependency. Dynamically build the local-temporal dependency simultaneously. This mechanisms can capture the factor wised dependency through density estimation with topology. The results shows the advance both in analytical and experimental. A further study may focus on the application of DualVDT and combining more informative differential equation model on specific fields.

\medskip

\printbibliography

\end{document}